\documentclass[10pt,twocolumn,letterpaper]{article}

\usepackage{iccv}
\usepackage{times}
\usepackage{epsfig}
\usepackage{graphicx}
\usepackage{amsmath}
\usepackage{amssymb}
\usepackage{color}
\usepackage{bbm}

\usepackage[breaklinks=true,bookmarks=false]{hyperref}

\iccvfinalcopy 

\def\iccvPaperID{****} 

\ificcvfinal\fi

\begin{document}

\title{Contrast R-CNN for Continual Learning in Object Detection}

\author{Kai Zheng\\
National University of Singapore\\
{\tt\small kzheng@u.nus.edu}
\and
Cen Chen\\
Institute for Infocomm Research, Singapore\\
{\tt\small chen\_cen@i2r.a-star.edu.sg}
}

\maketitle
\ificcvfinal\thispagestyle{empty}\fi

\begin{abstract}
   The continual learning problem has been widely studied in image classification, while rare work has been explored in object detection. Some recent works apply knowledge distillation to constrain the model to retain old knowledge, but this rigid constraint is detrimental for learning new knowledge. In our paper, we propose a new scheme for continual learning of object detection, namely Contrast R-CNN, an approach strikes a balance between retaining the old knowledge and learning the new knowledge. Furthermore, we design a Proposal Contrast to eliminate the ambiguity between old and new instance to make the continual learning more robust. Extensive evaluation on the PASCAL VOC dataset demonstrates the effectiveness of our approach.
\end{abstract}


\section{Introduction}
Object Detection is a general localization task to predict the bounding box and the category of the object in a scene. Recently, the development of CNN\cite{he2016deep, wang2020cspnet, tan2019efficientnet} gives rises to great advances in object detection\cite{cai2018cascade, zhu2019deformable, ren2015faster, lin2017focal, tian2019fcos, tan2020efficientdet}. However, the detector's learning way is not the same with normal training. One of the main gap is learning new objects in the dynamic world continually. Humans are born to learn continued knowledge while grasping the ability to keep the old knowledge. However, the state-of-the-art object detectors lacks the ability to imitate this continual learning paradigm. Although detectors can be fine-tuned for new tasks, they inevitably degrade the old knowledge while learn new tasks -- a problem called catastrophic forgetting\cite{goodfellow2013empirical, mccloskey1989catastrophic}.

In order to train an continual detector that can overcome the forgetting phenomenon, PASCAL VOC\cite{everingham2015pascal} and COCO\cite{lin2014microsoft} is used to build a benchmark for this new task. Due to the regulation of continual learning, the annotation of previous learned old object has to be removed and only the new object's annotation is kept in the dataset. Although the model pretrained on old object can be used for initialization on new task in the continual train step, the result is still unsatisfactory. The feature representation learned in old task can be drowned into the new task space, that is to say, the parameters learned before can be changed drastically on new task. To tackle this problem, continual learning of object detection is proposed to delve into the issue of catastrophic forgetting in detection task.

\begin{figure}[t]
\begin{center}
\resizebox{1.\linewidth}{!}{
\includegraphics[]{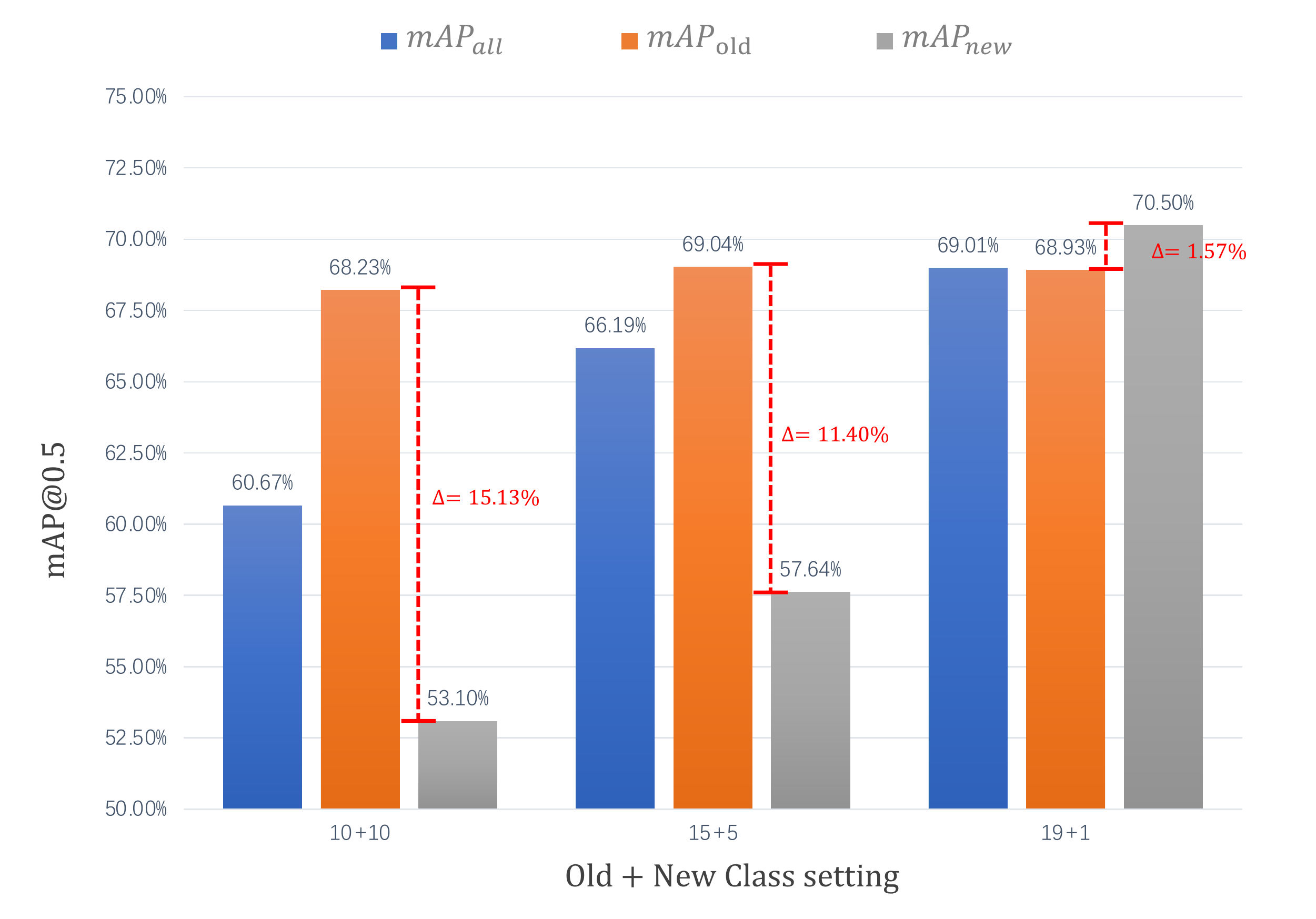}
}
\vspace{-2.5mm}
\caption{\textbf{The result of one distillation based continual detector.} We calculate the $\mathrm{mAP}$ over all categories, old categories and new categories respectively to show the unexpected phenomenon of the distillation based continual detector. In the figure, we observe an huge gap between the $\mathrm{mAP}_{old}$ and $\mathrm{mAP}_{new}$. We denotes this phenomenon as \textit{imbalanced performance} in distillation based method.}
\label{fig:motivation}
\vspace{-5.5mm}
\end{center}
\end{figure}

\begin{figure}[t]
\begin{center}
\resizebox{0.95\linewidth}{!}{
\includegraphics[]{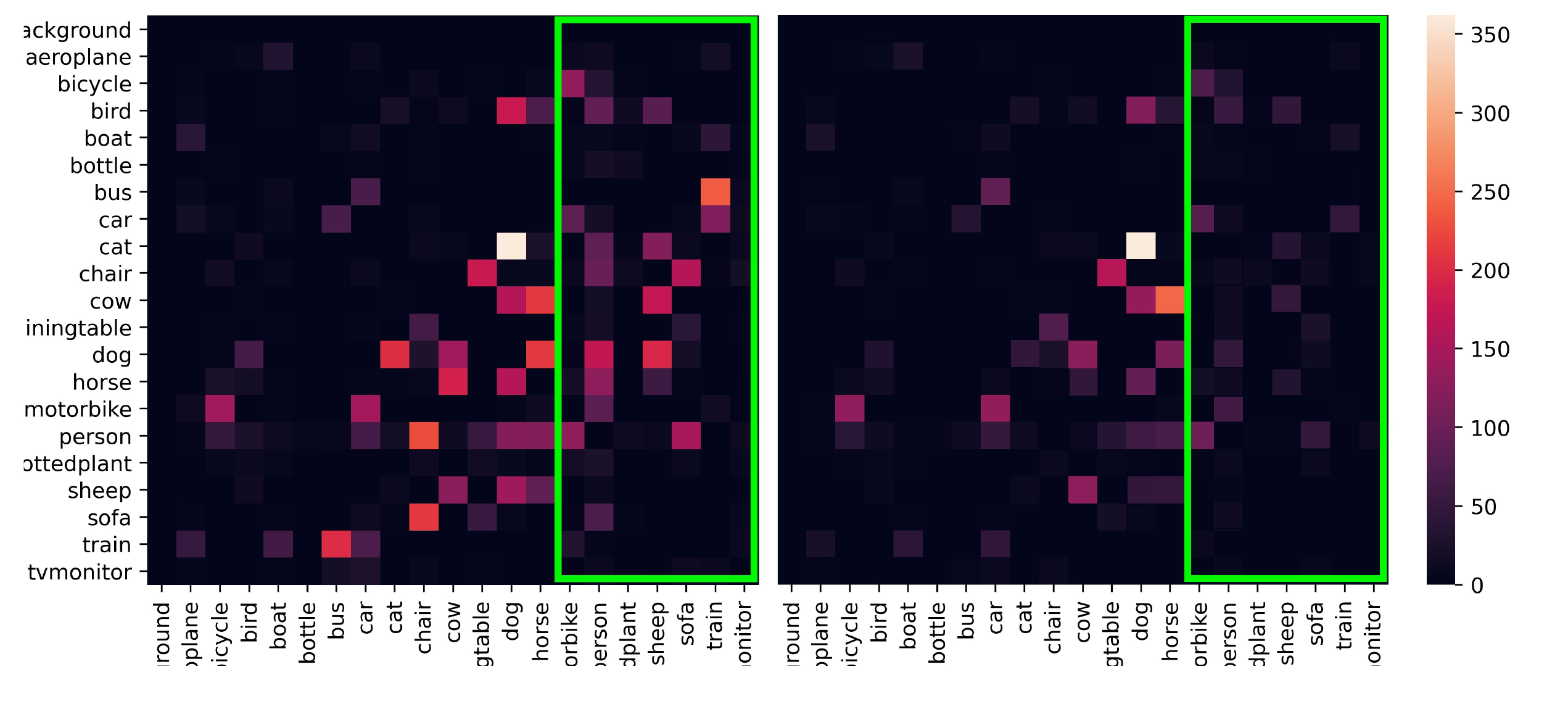}
}
\vspace{1.5mm}
\caption{\textbf{Confusion Matrix for all categories for misclassification illustration.} We select 15+5 setting for illustration, \eg first fifteen categories as old classes, the rest five categories as new classes. The horizontal axis and vertical axis stand for the prediction category and ground-truth category respectively. We annotate a \textcolor[RGB]{10,123,58}{green} box for better visualization of the old classes that misclassified to the new classes. The \textit{left} figure stands for the Faster ILOD (distillation-based method) while the \textit{right} figure represents our method, \ie Contrast R-CNN. We observe that our Contrast R-CNN can alleviate the misclassification effect between old and new category. Note, the diagonal cell's value is ignored when visualization for better analysis.}
\label{fig:motivation2}
\vspace{-6.5mm}
\end{center}
\end{figure}

Existing continual learning object detectors\cite{shmelkov2017incremental, hao2019end, peng2020faster} use distillation to mitigate forgetting. These methods requires the model trained on old dataset to generate the constraints to regularize the weight updates. The constraints can be obtained from either feature-level or response-level, which denotes the feature map regularization and classification distribution regularization respectively. However, these distillation-based method has some critical deficiency, that is \textit{imbalanced performance} and \textit{confusion classification}. In Fig.\ref{fig:motivation}, we show the Faster ILOD's (one of the typical distillation based method) performance on three datset splitting with regard to three $\mathrm{mAP}_{all}, \mathrm{mAP}_{old}, \mathrm{mAP}_{new}$ metrics. We can observe that there exists a huge gap between the performance on old and new categories set. This implies the distillation method's good performance on old categories is at the expense of performance on new categories. We also delve into the distillation based method and find that if we re-balance the loss weight between the original loss and distillation loss, the huge gap on old and new categories can be alleviate. This findings verifies our hypothesis that the distillation put too heavy constraints on the detector to retain the old knowledge, which prevents from learning the new knowledge. Another \textit{confusion classification} problem is the misclassification when performing classifier on objects. From Fig.\ref{fig:motivation2} (left), we can observe that the misclassification exists between old and new categories, and similar categories. To some extent, we can attribute this misclassification problem to the distillation paradigm. The distillation based method place too strict constraints on old knowledge so that the detector minimizes the loss by imitating all the old knowledge rather than learning the new knowledge. We believe that without further constraints on discrimination between old and new classes, the model can be drowned into the classification easily.

A common approach to solve the confusion problem is to use a margin classifier~\cite{elsayed2018large} or apply contrastive learning~\cite{chopra2005learning}. The exploration of self-supervise learning has also arouse researchers' interest in contrastive learning. In self-supervised contrastive learning for image classification\cite{mocov2, chen2020simple}, image augmentation is utilized to build the positive pairs of two views from one image while the negative pairs are those from different images.\cite{khosla2020supervised} extend the contrastive learning into the supervise paradigm, where the images from the same class are used to enrich the positive pairs. We think the contrastive learning can help alleviate the confusion problem among classes. In detection task, the label assignment can generate the semantic category for each proposal, which can be used to build positive pairs in proposal-level. Therefore, in our work, we explore the supervised contrastive learning in continual detection. We believe the representation learned contrastively will enlarge the decision boundary among different classes and ease the misclassification.

We present Contrast R-CNN, a better approach for continual learning of object detection. To build the contrastive objective, we need to obtain adequate proposals from different categories, including old ones. However, the old instances is missing in dataset due to the continual setting. Inspired by some semi-supervised detection methods\cite{bachman2014learning, lee2013pseudo}, we propose an entropy-based data distillation to obtain the ``old instance''. With the proposals labeled by ground-truth and distilled data, we can leverage the supervised contrastive learning to do contrast based on categories. 

To the best of our knowledge, we. are the first to bring contrastive learning into continual learning of object detection. We benchmark Contrast R-CNN with continual setting using the PASCAL VOC datasets under different experiment setting, namely 10+10, 15+5 and 19+1. Our detector sets the new state-of-the-art in two benchmarks, with up to +4.36\% and +0.61\% on all category and almost +10\% on new category in 10+10 and 15+5 settings.

\begin{figure*}[h]
\begin{center}
\resizebox{0.9\linewidth}{!}{
\includegraphics[]{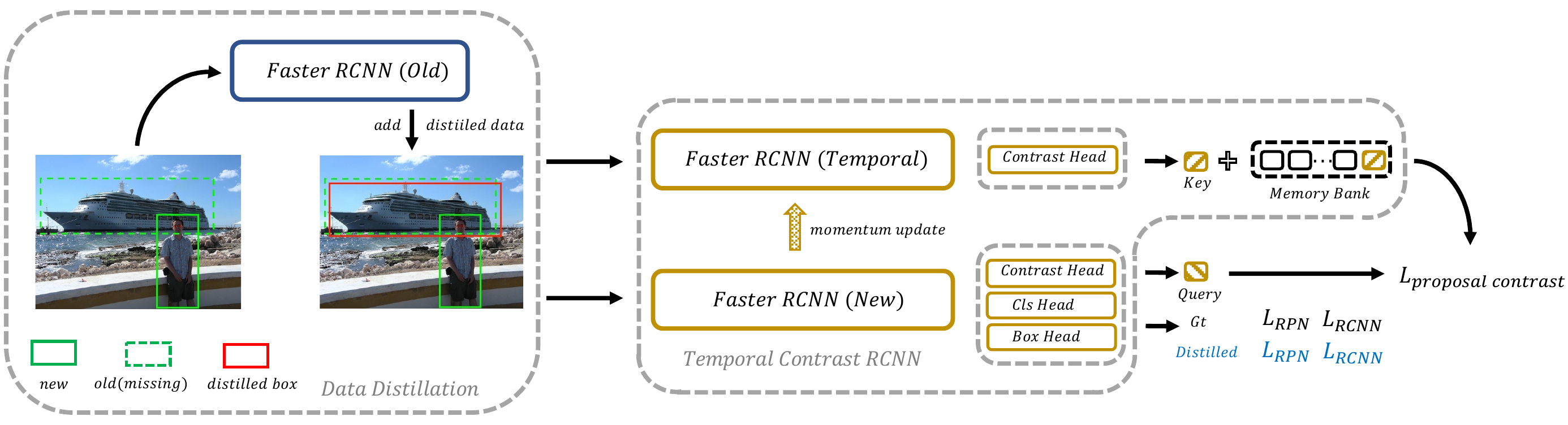}
}
\vspace{1.5mm}
\caption{\textbf{Overview of Contrast R-CNN.} Our framework mainly consists of two parts, Data Distillation and Temporal Contrast R-CNN. The Data Distillation is used to distill the label knowledge for continual training. To learn the discriminative representation, we introduce a novel contrastive sub-framework.}
\label{fig:overview}
\vspace{-6.0mm}
\end{center}
\end{figure*}

\section{Related Work}
\noindent{\textbf{Continual Learning.}}
Continual Learning refers to learn a new knowledge continuously without losing the ability to retain the knowledge learned previously.\cite{li2017learning} uses the expanded classifier to extract the new knowledge and applies cross-entropy and knowledge distillation loss on new class and old class respectively.\cite{rebuffi2017icarl} stores some old data as exemplar and combine these old data with new data to train the new model.\cite{castro2018end} retains all the classifier for distill knowledge on new task.\cite{wu2019large} proposes to form a balanced old data and new data for each batch to train the offset for alleviate the bias on new class. In contrast to the continual learning of classification, rare work has been explored in continual learning of object detection.\cite{shmelkov2017incremental} proposes to apply distillation on proposals generated by external proposal generator.\cite{hao2019end, peng2020faster} both leverage the knowledge distillation on either the form of feature, response knowledge or altogether.\cite{li2019rilod} also applies the distillation, but uses other one-stage object detector, \eg RetinaNet. All these continual learning methods are based on distillation, which is the simple but effective way to prevent from forgetting. However, in distillation based continual learning methods, the new trained detector relies heavily on the previous old trained model, in other words, the new model just imitate the response on old class, not considering the relation between old and new class. In our experiments, we find that the distillation based method puts too strict constraints on old class and results in the bad performance on new class. In contrast to the previous methods, we target designing an continual detector performing balanced on both old and new class.

\noindent{\textbf{Contrastive Learning.}}
The recent success of self-supervised methods has validated the effectiveness of contrastive learning.\cite{oord2018representation, wu2018unsupervised, he2020momentum, mocov2, chen2020simple} formulate the way of contrastive learning as contrasting between positive and negative pairs. The objective is to maximize the agreement between semantic similar instances and differentiate the dissimilar.\cite{khosla2020supervised} extends the contrastive learning into supervise learning, and the result is appealing. To our knowledge, our work is the first to investigate contrastive learning in continual learning of object detection.

\noindent{\textbf{Semi-Supervised Learning.}}
Semi-supervised learning explores the way of exploiting unlabeled data to facilitate the model training of on limited annotated data. Pseudo labeling\cite{bachman2014learning, lee2013pseudo}, one of the simple and natural way to extract the extra information from the unlabeled data, is one of the popular class: the model generates pseudo labels on unlabeled data and thereby trained a new model with this augmented data with labeled data.\cite{xie2020self} proposes an iterative teacher-student strategy that redoes the labeling assignments using the teacher model and train the student model.\cite{sohn2020simple} pretrains a detector using the labeled data and uses it to generate pseudo labels on unlabeled data to fine-tune the pretrained model with full data.\cite{zhou2021instant} proposes an online pseudo labeling strategy to keep the model that generates pseudo labels updating with train. In continual learning of object detection, we aim to explore the potential of leveraging the previous trained model in continual learning to benefit the new model.

\section{Problem Definition and Setups}

Continual learning aims at learning a model in $t=1...T$ steps. In each continual step, we present a dataset $D_t$ that consists of a set of pairs $\{(x_n,y_n)\}_{n=1}^N$, where $x_n$ denotes the input image and $y_n$ denotes the corresponding ground truth bounding box and category annotation. The latter only contains the labels of current class $C^t$, and all the other labels (\eg old categories $C^{1:t-1}$ or future categories $C^{t+1:T}$) is unaccessible. In the data setting, these bounding box annotations are removed. However, the model at step $t$ is required to predict all the classes seen before, \eg $C^{1:t}$. In most continual learning\cite{shmelkov2017incremental, peng2020faster} setting, $T$ is set to 2, that is to say, $C^1$ and $C^2$ refer to old categories $C_{old}$ and new categories $C_{new}$ respectively.

\section{Method}
Our proposed Contrast R-CNN involves a simple two-stage training. First, a standard Faster R-CNN is trained on old data ($D_{train}=D_{old}$). Then, the old detector is transferred to new data through fine-tuning on new dataset with the old instance annotation removed ($D_{train}=D_{new}$).

In the following section, we show an overview of our Contrast R-CNN framework (see Section \ref{Method:Overview of Framework}), which consists of a Data Distillation module (see Section \ref{Method:Pseudo Labeling}) and a Temporal Contrast R-CNN (see Section \ref{Method:Temporal Contrast RCNN}).

\subsection{The overview of framework}\label{Method:Overview of Framework}

\begin{figure}[b]
\begin{center}
\resizebox{0.9\linewidth}{!}{
\includegraphics[]{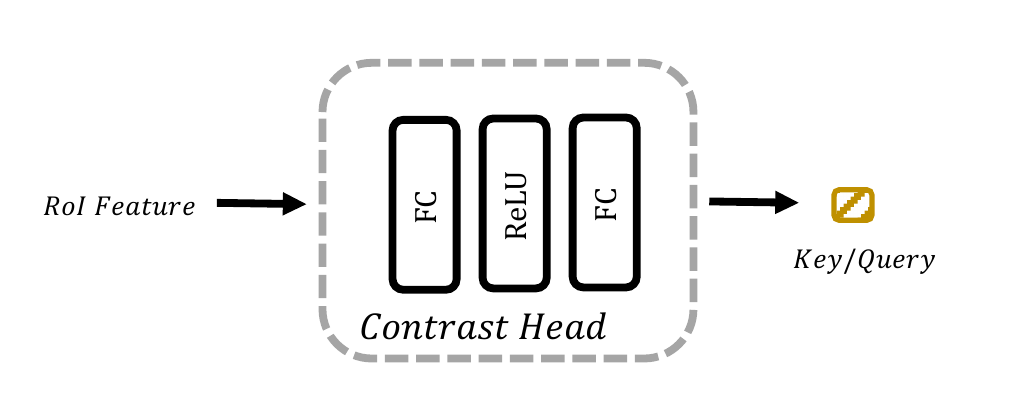}
}
\vspace{-2.5mm}
\caption{\textbf{Illustration of Contrast Head.} The contrast is composed of a $\mathrm{FC-ReLU-FC}$ structure to map the roi feature into the embedding space. Two $\mathrm{FC}$s are learnable in training.}
\label{fig:contrasthead}
\vspace{-4.5mm}
\end{center}
\end{figure}

As shown in Fig.\ref{fig:overview}, our Contrast R-CNN framework is mainly composed of two modules, namely, Data Distillation and Temporal Contrast R-CNN. In Data Distillation, the well-trained Faster RCNN processes the input image and distills the box based on the old knowledge learned in previous task. After getting the raw distilled box, we apply the \textit{median entropy filter} to process the distilled box. Only the relative confident instances will be kept as final distilled box. Then distilled box (old instance) and ground truth box (new instance) will be fed into the next step as supervision. In Temporal Contrast RCNN, it consists of two same Faster RCNN, namely Temporal and New, but with different heads. For New Faster RCNN, a Contrast Head (as illustrated in Fig.~\ref{fig:contrasthead}) that mapping the roi feature into the embedding space is added parallelly to the Cls Head and Box Head. For Temporal Faster RCNN, only Contrast Head is kept. After getting the embeddings from Temporal and New Faster RCNN, namely Key and Query, a supervised proposal contrast objective is employed to pull the same category proposal together and push different category object apart in embedding space. In parallel to the contrastive objective, the classification and localization objectives are maintained under the distilled plus ground-truth fashion.

\subsection{Data Distillation}\label{Method:Pseudo Labeling}

\noindent{\textbf{Data Selection.}}
In semi-supervised object detection, previously trained model is used to extract the knowledge in unlabeled data to provide extra information. Likewise, in continual learning, we exploit the old model to mine the useful knowledge in the new data. In our method, we propose a data distillation module to distill the old knowledge in the form of box prediction to benefit the new model training. We find that it is unreasonable to apply a simple threshold for all the old classes, because different category has distinctive distribution in the model prediction. Therefore, we introduce a \textit{median entropy filter} strategy. For each distilled bounding box $B_i$, the strategy computes an entropy $e_i$ for the classification distribution alongside its prediction category $c_i$. Each category $C_j$ will form an entropy set $E_{C_j}=\{e_k | c_k = C_j, k=1,...,N\}$, and compute the median value in the set $E_j$ as the threshold $\alpha_{C_j}$ of this category. After getting the threshold for each category, we apply the category-specific median threshold filter on the raw distilled box, only the box with the entropy lower than the median threshold will be kept.

\noindent{\textbf{Data Uncertainty.}}
It is intuitive that the distilled box and ground-truth box can not be equivalent due to the inherent noisy in prediction. Therefore, we consider the entropy for each distilled box as the uncertainty weight in loss calculation. For each box ($B_i$, $c_i$) with the entropy $e_i$, the uncertainty weight $w_i$ can be computed as:

\begin{equation}\label{eq1}
\resizebox{0.6\linewidth}{!}{
$w_i=\frac{\mathrm{max}\{E_{c_i}\} - e_i}{\mathrm{max}\{E_{c_i}\}-\mathrm{min}\{E_{c_i}\}}$
}
\end{equation}

where,

\begin{equation}\label{eq7}
\resizebox{0.43\linewidth}{!}{
$e_i=-\sum_{j=0}^C p_i^j\mathrm{log}p_i^j$
}
\end{equation}

$c_i$ is the $i$-th bounding box's prediction category, $C$ is the the summation of $C_{old}$ and $C_{new}$, $p_i$ is the classification probability of for $i$-th bounding box.

In Faster RCNN, it sets a Cls Head and a Box Head to predict each RoI's classification score and box offset. For our Contrast R-CNN, we maintain these two heads and compute the cross entropy loss and smooth L1 loss respectively. The uncertainty weight is considered to re-scale the distilled box loss. The weight is applied in both $L_{RPN}$ and $L_{RCNN}$ calculation. In our experiments, we find that this uncertainty re-weighting strategy will make the model robust to false prediction.

\subsection{Temporal Contrast R-CNN}\label{Method:Temporal Contrast RCNN}
\noindent{\textbf{Proposal Contrast Embedding.}}
Despite the introduction of data distillation as the complement for ground-truth annotation, the continual detector is still unable to learn a clear decision boundary because of the unconfident distilled data. This leads to the confusion between old and new category or misclassification between similar categories. To obtain more robust feature representation from limited accurate annotation, our idea is to make the model learn discriminative proposal embedding in continual learning stage.

To this end, we design a Temporal Contrast R-CNN with a Memory Bank for enriching the discriminative feature. The Temporal Contrast R-CNN incorporates two Faster RCNN, a normal Faster RCNN (New) and a Faster RCNN (Temporal). They share the same feature extractor but with different heads. In order to map the RoI feature into the embedding space, we introduce a Contrast Head parallel to classification and box head. In the embedding space, the semantic distance can be measured with angle.

In order for learning the discriminative feature, we set a contrastive objective, which contrasts positive pairs against negative pairs\cite{mocov2, he2020momentum, chen2020simple}. To provide more inter-class diversity, we establish a Memory Bank to store the diverse negative pairs. The Memory Bank can be regarded as an FIFO queue, where embeddings (Keys) of the current mini-batch from Faster RCNN (Temporal) are enqueued, and the oldest are dequeued. The length of queue decouples from the mini-batch size, allowing it to be large. Moreover, the embeddings from the Faster RCNN (Temporal) are ensured to be feature consistency following the special parameter update scheme\cite{he2020momentum}. The Faster RCNN (Temporal)'s parameters are exponential momentum updated from the gradient updated Faster RCNN (New):

\begin{equation}\label{eq2}
\resizebox{0.4\linewidth}{!}{
$\theta_t \leftarrow m\theta_t + (1-m)\theta_n$
}
\end{equation}

Here, $m\in [0,1)$ is a momentum coefficient, which is used to control the scale of evolving. $\theta_t, \theta_n$ denote the parameters of Faster RCNN (Temporal) and Faster RCNN (New) respectively. Only the parameters $\theta_n$ are updated by back-propagation. In experiments, a relatively large momentum (\eg $m=0.999$, our default) ensures the $\theta_t$ evolve smoothly and slowly.

\begin{table*}[h]
\begin{center}
\resizebox{0.8\linewidth}{!}
{
\begin{tabular}{lll|ccc|ccc|ccc}
\hline
\multicolumn{3}{c|}{} & \multicolumn{3}{c|}{\bf{10+10 setting}} & \multicolumn{3}{c|}{\bf{15+5 setting}} & \multicolumn{3}{c}{\bf{19+1 setting}} \\
\multicolumn{1}{r}{} &  &  & $\mathrm{mAP}_{old}$  &  $\mathrm{mAP}_{new}$ & $\mathrm{mAP}$ & $\mathrm{mAP}_{old}$  &  $\mathrm{mAP}_{new}$ & $\mathrm{mAP}$ & $\mathrm{mAP}_{old}$  &  $\mathrm{mAP}_{new}$  &  $\mathrm{mAP}$    \\ \hline
Normal R-CNN (fine-tune) & & & 13.88 & 62.50 & 38.19 & 18.17 & 56.14 & 27.66 & 5.74 & 78.94 & 9.40  \\ \hline
Faster ILOD\cite{peng2020faster} & & & 68.23 & 53.10 & 60.67 & 69.04 & 57.64 & 66.19 & 68.93 & 70.50 & 69.01 \\ \hline
Contrast R-CNN (ours)  & & & 66.62 & \textbf{63.44} & \textbf{65.03} & 66.82 & \textbf{67.23} & \textbf{66.92} & 68.02 & 67.82 & 68.01 \\ \hline
\end{tabular}
}
\end{center}
\vspace{-1.2mm}
\caption{Results on three different benchmark settings. We use $\mathrm{mAP}_{old}$, $\mathrm{mAP}_{new}$, $\mathrm{mAP}$ as metrics to evaluate the performance on old, new and overall categories. Normal R-CNN (fine-tune) refers to fine-tuning on the new data simply. The Faster ILOD's result is reproduced by us based on the publicly released code.}
\vspace{-1.2mm}
\label{table:table1}
\end{table*}

\begin{table*}[h]
\begin{center}
\resizebox{0.9\linewidth}{!}
{
\begin{tabular}{lll|ccc|ccc|ccc}
\hline
\multicolumn{3}{c|}{} & \multicolumn{3}{c|}{\bf{10+10 setting}} & \multicolumn{3}{c|}{\bf{15+5 setting}} & \multicolumn{3}{c}{\bf{19+1 setting}} \\
\multicolumn{1}{r}{} &  &  & $\mathrm{mAP}_{old}$  &  $\mathrm{mAP}_{new}$ & $\mathrm{mAP}$ & $\mathrm{mAP}_{old}$  &  $\mathrm{mAP}_{new}$ & $\mathrm{mAP}$ & $\mathrm{mAP}_{old}$  &  $\mathrm{mAP}_{new}$  &  $\mathrm{mAP}$    \\ \hline
Contrast R-CNN w/o memory bank  & & & 63.31 & 64.34 & 63.83 & 65.35 & 67.97 & 66.00 & 64.26 & 80.06 & 65.05 \\ \hline
Contrast R-CNN w/o median selection  & & & 65.61 & 63.73 & 64.67 & 66.64 & 67.22 & 66.78 & 67.85 & 66.45 & 67.78 \\ \hline
Contrast R-CNN w/ memory bank, median selection  & & & \textbf{66.62} & 63.44 & \textbf{65.03} & \textbf{66.82} & 67.23 & \textbf{66.92} & \textbf{68.02} & 67.82 & \textbf{68.01} \\ \hline
\end{tabular}
}
\end{center}
\vspace{-3.2mm}
\caption{Ablation Study. Three rows refer to the method proposed by us that without memory bank, without median selection, with memory bank and median selection respectively.}
\vspace{-2.2mm}
\label{table:table2}
\end{table*}

\noindent{\textbf{Proposal Contrastive Loss.}}
For the contrastive objective, we construct the embeddings from Faster RCNN (New) and embeddings with the same category label in Memory Bank as positive pairs, while others as negative pairs. Then we formulate our Proposal Contrastive Loss $L_{PC}^i$ as follows:

\begin{equation}\label{eq3}
\resizebox{0.98\linewidth}{!}{
$L_{PC}^i = -\frac{1}{N_{c_i}} \sum_{j=1,j\neq i}^N \mathbbm{1}\{c_i=c_j\}\cdot log\frac{\mathrm{exp}(z_i\cdot z_j/\tau)}{\sum_{k=1}^N \mathbbm{1}_{k\neq i}\cdot \mathrm{exp}(z_i\cdot z_k)/\tau}$
}
\end{equation}

$N_{c_i}$ is the number of embeddings with the same label $c_i$ in Memory Bank, $N$ is the total number of embeddings in Memory Bank, and $\tau$ is the hyper-parameter temperature as in InfoNCE\cite{oord2018representation}.

There exists the uncertainty variance among the proposals, and the reason mainly lies in two-fold: (i) the uncertainty in roi-distilled box's matching. The lower the IoU is, the more the uncertainty is. (ii) the uncertainty in distilled box itself. The latter one comes from the pretrained model's noise. To ameliorate the effect of uncertainty, we introduce two variables $\phi_1, \phi_2$ to model each uncertainty respectively:

\begin{equation}\label{eq4}
\resizebox{0.32\linewidth}{!}{
$\phi_1^{i,j}=IoU(P_i,R_j)$
}
\end{equation}

\begin{equation}\label{eq5}
\resizebox{0.17\linewidth}{!}{
$\phi_2^i=w_i$
}
\end{equation}

where $P_i,R_j$ represent the $i$-th distilled box and $j$-th roi. The calculation for $w_i$ refers to Eq.\ref{eq1}. Then, the $L_{PC}^i$ is reformulated as

\begin{equation}\label{eq6}
\resizebox{0.99\linewidth}{!}{
$L_{PC}^i = -\frac{1}{N_{c_i}} \cdot \phi_2^i \cdot \sum_{j=1,j\neq i}^N \mathbbm{1}\{c_i=c_j\}\cdot \phi_1^{i,j} log\frac{\mathrm{exp}(z_i\cdot z_j/\tau)}{\sum_{k=1}^N \mathbbm{1}_{k\neq i}\cdot \mathrm{exp}(z_i\cdot z_k)/\tau}$
}
\end{equation}

\section{Experiments}

\noindent{\textbf{Implement Details.}}
The results for Faster ILOD\cite{peng2020faster} are generated using their public implementation. For fair comparison, we reproduce the result under the same experiment setting. We use the same backbone network (ResNet-50) and similar training strategy stated in their paper. In the first step of training an old model, we set the learning rate to 0.001, decaying to 0.0001 at 30k iterations, weight decay is set to 0.0001 and the momentum is 0.9. The total training iteration is 40k for PASCAL VOC. In the following continual training step, learning rate is set to 0.001, other hyper-parameters keeps the same as the first step.

\noindent{\textbf{Dataset and Evaluation.}}
We evaluate our method on PASCAL VOC 2007. VOC 2007 is composed of 5K images in the trainval set and 5K images in the test set of 20 object categoires. For the evaluation metric, we use the standard mean average precision (mAP) at 0.5 Intersection over Union (IoU). Evaluation of the VOC 2007 is done on test set. Table \ref{table:table1} shows the results for continual setting, where 10+10, 15+5, 19+1 represents the old+new category split (\eg 10+10 refers to first ten categories for old first step training, the rest ten categories for continual step training). In order to delve into the performance on old and new category, we calculate the mAP on old and new categories respectively, denoted by $\mathrm{mAP}_{old}$, $\mathrm{mAP}_{new}$. We observe that in 10+10 and 15+5 setting, our Contrast R-CNN has achieved SOTA in both $\mathrm{mAP}$ and $\mathrm{mAP}_{new}$. Especially in $\mathrm{mAP}_{new}$, we outperforms Faster ILOD a large margin by almost +10\%. This can be attributed to the fact that the distillation-based method's bad performance on new class. From the Table \ref{table:table1}, it is obvious that the huge gap exists between $\mathrm{mAP}_{old}$ and $\mathrm{mAP}_{new}$ on all three settings. This fact convinces us that the distillation pose too strict constraints on continual model to retain the old knowledge, which inevitably hold back the learning of new knowledge. This spectulation exactly help interpret the more old categoires are, the better its performance is, that's why Faster-ILOD performs slightly better than us in 19+1 setting. Meanwhile, this severe imbalanced performance between the old and new knowledge reveals the critical deficiency in distillation-based method in continual learning, which is ignored by most researchers\cite{peng2020faster, hao2019end, li2019rilod}. In contrast, our method shows an reasonable balanced performance on both knowledge.

\section{Ablation}
To evaluate the contributions of each component, we conduct a series of experiments and report the ablation analysis in Table ~\ref{table:table2} on PASCAL VOC dataset in 10+10, 15+5, 19+1 settings. As already noticed, Contrast R-CNN without the memory bank leads to a great degradation, where $\mathrm{mAP}_{old}/\mathrm{mAP}$ suffers from -3.31\%/-1.2\%, -1.47\%/-0.92\%, -3.76\%/-2.96\% in three settings. The median selection strategy can help improve the accuracy of old categories slightly.

\section{Conclusion}
In this paper, we paved a new way for future research on Continual Learning for Object Detection, which is an emerging and practical domain in computer vision. In particular, our Contrast R-CNN demonstrates the new perspective in dealing with the missing annotation and saving the object from being misclassified. Our method achieves state-of-the-art results in most settings. For a broader impact, our work proves the plausibility of incorporating the contrastive learning into the continual object detection. We hope our work can give more inspiration for researchers.


\bibliographystyle{ieee_fullname}
\bibliography{egbib}

\end{document}